\documentclass[11pt,a4paper]{article}
\usepackage[hyperref]{acl2021}
\usepackage{times}
\usepackage{latexsym}

\usepackage{graphicx}
\usepackage{amsmath}
\usepackage{amsfonts}
\usepackage{bbm}
\usepackage{amssymb}
\usepackage{booktabs}
\usepackage{multirow}
\usepackage{paralist}
\usepackage{microtype}
\aclfinalcopy % Uncomment this line for the final submission
\def\aclpaperid{1016} %  Enter the acl Paper ID here

\title{\textsc{One2Set}: Generating Diverse Keyphrases as a Set}

\author{
Jiacheng Ye$^{1}$, 
Tao Gui$^{2}$\thanks{$^*$ Corresponding authors.}, 
Yichao Luo$^{1}$, 
Yige Xu$^{1}$ and 
Qi Zhang$^{1*}$ \\
$^1$School of Computer Science, Fudan University \\
$^2$Institute of Modern Languages and Linguistics, Fudan University \\
{\tt \{yejc19, tgui16, ycluo18, ygxu18, qz\}@fudan.edu.cn}
}

\date{}
\begin{document}
\maketitle
\begin{abstract}
Recently, the sequence-to-sequence models have made remarkable progress on the task of keyphrase generation (KG) by concatenating multiple keyphrases in a predefined order as a target sequence during training.
However, the keyphrases are inherently an unordered set rather than an ordered sequence.  
Imposing a predefined order will introduce wrong bias during training, which can highly penalize shifts in the order between keyphrases. 
In this work, we propose a new training paradigm \textsc{One2Set} without predefining an order to concatenate the keyphrases. To fit this paradigm, we propose a novel model that utilizes a fixed set of learned control codes as conditions to generate a set of keyphrases in parallel. 
To solve the problem that there is no correspondence between each prediction and target during training, we propose a $K$-step target assignment mechanism via bipartite matching, which greatly increases the diversity and reduces the duplication ratio of generated keyphrases.
The experimental results on multiple benchmarks demonstrate that our approach significantly outperforms the state-of-the-art methods.
\end{abstract}

\section{Introduction}
Keyphrase generation (KG) aims to generate of a set of keyphrases that expresses the high-level semantic meaning of a document. These keyphrases can be further categorized into \textit{present} keyphrases that appear in the document and \textit{absent} keyphrases that do not. 
\citet{meng2017} proposed a sequence-to-sequence (Seq2Seq) model with a copy mechanism \citep{gu2016Incorporating} to predict both present and absent keyphrases. However, the model needs beam search during inference to overgenerate multiple keyphrases, which cannot determine the dynamic number of keyphrases. To address this, \citet{yuan2018} proposed the \textsc{One2Seq} training paradigm where each source text corresponds to a sequence of keyphrases that are concatenated with a delimiter $\langle\mathit{sep}\rangle$ and a terminator $\langle\mathit{eos}\rangle$. As keyphrases must be ordered before being concatenated, \citet{yuan2018} sorted the present keyphrases by their order of the first occurrence in the source text and appended the absent keyphrases to the end. During inference, the decoding process terminates when generating $\langle\mathit{eos}\rangle$, and the final keyphrase predictions are obtained after splitting the sequence by $\langle\mathit{sep}\rangle$. Thus, a model trained with \textsc{One2Seq} paradigm can generate a sequence of multiple keyphrases with dynamic numbers as well as considering the dependency between keyphrases.

\begin{figure}[t]
\centering
\includegraphics[width=2.7in]{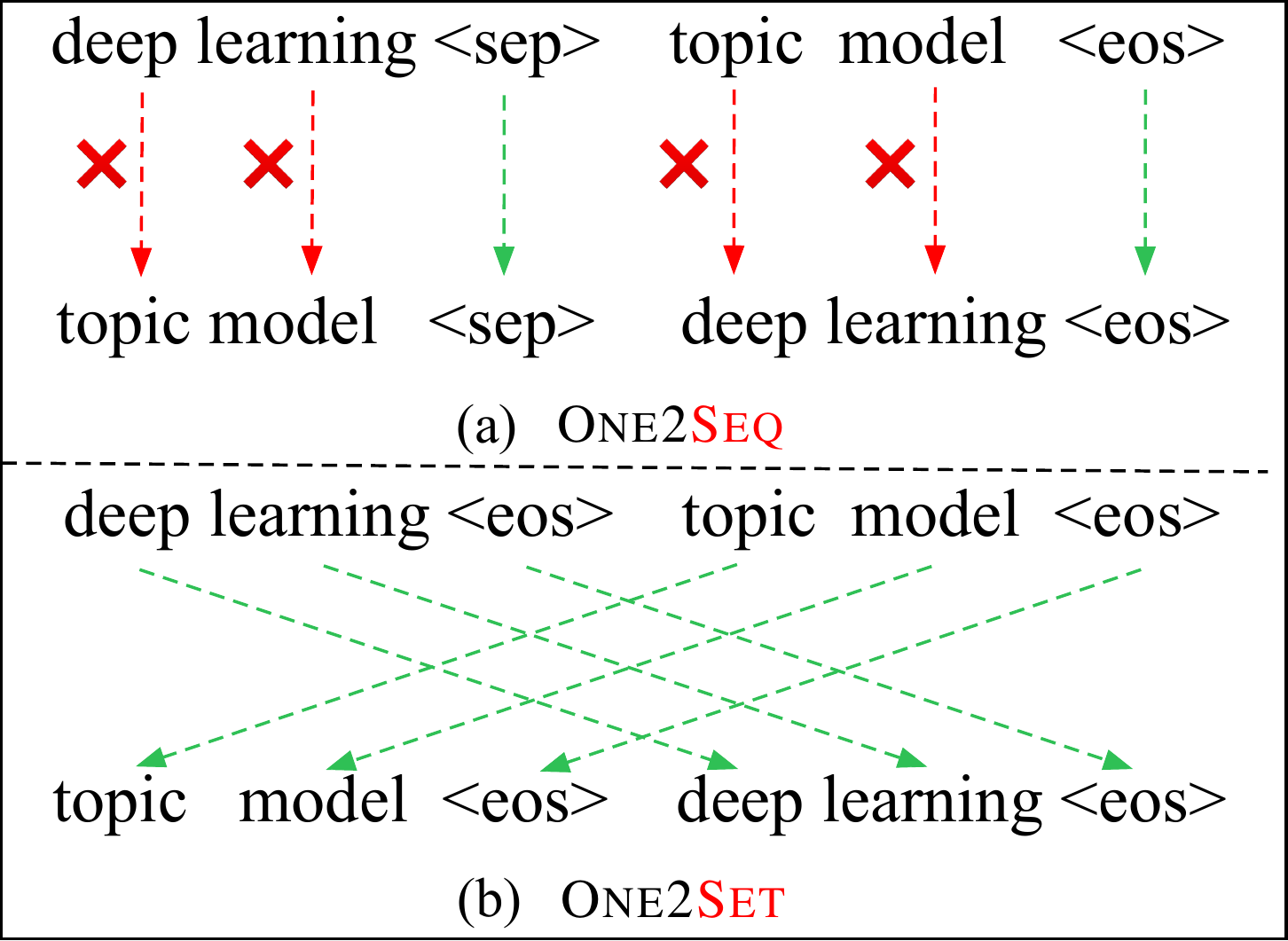}
\caption{An example of ground-truth keyphrases (upper) and predictions (lower) under \textsc{One2Seq} and \textsc{One2Set} training paradigm. For the \textsc{One2Seq} training paradigm, although the predictions are correct in each keyphrase, they will still be considered wrong  due to the shift in keyphrase order, and the model will receive a large penalty.}
\label{fig:motivation}
\end{figure}

However, as the keyphrases are inherently an unordered set rather than an ordered sequence, imposing a predefined order usually leads to the following intractable problems. 
First, the predefined order will give wrong bias during training, which can highly penalize shifts in the order between keyphrases. As shown in Figure \ref{fig:motivation} (a), the model makes correct predictions in each keyphrase but can still receive a large loss during training.
Second, this increases the difficulty of model training. For example, the absent keyphrases are appended to the end in an author-defined order in \citet{yuan2018}, however, different authors can have various sorting bases, which makes it difficult for the model to learn a unified pattern. 
Third, the model is highly sensitive to the predefined order, as shown in \citet{meng2019}, and can suffer from error propagation during inference when previously having generated keyphrases with an incorrect order. 
Lately, \citet{chan2019} proposed a reinforcement learning-based fine-tuning method, which fine-tunes the pre-trained models with metric-based rewards (i.e., recall and $F_1$) for generating more sufficient and accurate keyphrases. However, this method can alleviate the impact of the order problems when fine-tuning but needs to be pre-trained under the \textsc{One2Seq} paradigm to initialize the model, which can still introduce wrong biases.

To address this problem, we propose a new training paradigm \textsc{One2Set} where the ground-truth target is a set rather than a keyphrase-concatenated sequence. 
However, the vanilla Seq2Seq model can generate a sequence but not a set. Hence, we introduce a set prediction model that adopts Transformer \citep{vaswani2017attention} as the main architecture together with a fixed set of learned control codes as additional decoder inputs to perform controllable generation. 
For each code, the model generates a corresponding keyphrase for the source document or a special $\varnothing$ token that represents the meaning of ``no corresponding keyphrase''. During training, the cross-entropy loss cannot be directly used since we do not know the correspondence between each prediction and target. Hence, we introduce a $K$-step target assignment mechanism, where we first auto-regressively generate $K$ words for each code and then assign targets via bipartite matching based on the predicted words. After that, we can train each code using teacher forcing as before. 
Compared with the previous models, the proposed method has the following advantages: 
(a) there is no need to predefine an order to concatenate the keyphrases, thus the model will not be affected by the wrong biases in the whole training stage; and (b) the bipartite matching forces unique predictions for each code, which greatly reduces the duplication ratio and increases the diversity of predictions.

We summarize our main contributions as follows: 
(1) we propose a new training paradigm \textsc{One2Set} without predefining an order to concatenate the keyphrases;
(2) we propose a novel set prediction model that can generate a set of diverse keyphrases in parallel and a dynamic target assignment mechanism to solve the intractable training problem under the \textsc{One2Set} paradigm;
(3) our method consistently outperforms all the state-of-the-art methods and greatly reduces the duplication ratio. 
Our codes are publicly available at \textit{Github}\footnote{https://github.com/jiacheng-ye/kg\_one2set}.

\section{Related Work}
\subsection{Keyphrase Extraction}
Existing approaches for keyphrase prediction can be broadly divided into extraction and generation methods. 
Early work mostly focuses on the keyphrase extraction task, and a two-step strategy is typically designed
 \citep{hulth2003improved,mihalcea2004textrank,nguyen2007keyphrase,wan2008single}. First, they extract a large set of candidate phrases by hand-crafted rules \citep{mihalcea2004textrank,medelyan2009human,liu2011automatic}. Then, these candidates are scored and reranked based on unsupervised methods \citep{mihalcea2004textrank,wan2008single} or supervised methods \citep{hulth2003improved,nguyen2007keyphrase}. 
Other extractive approaches utilize neural-based sequence labeling methods \citep{zhang2016keyphrase,gollapalli2017incorporating}. 

\subsection{Keyphrase Generation}
Compared to extractive approaches, generative ones have the ability to consider the absent keyphrase prediction. \citet{meng2017} proposed a generative model CopyRNN, which employs a encoder-decoder framework \citep{sutskever2014sequence} with attention \citep{bahdanau2014neural} and copy mechanisms \citep{gu2016Incorporating}. Many works are proposed based on the CopyRNN architecture \citep{chen2018a,zhao2019,chen2019a,chen2019}. 
\begin{figure*}[t]
\centering
\includegraphics[width=5.2in]{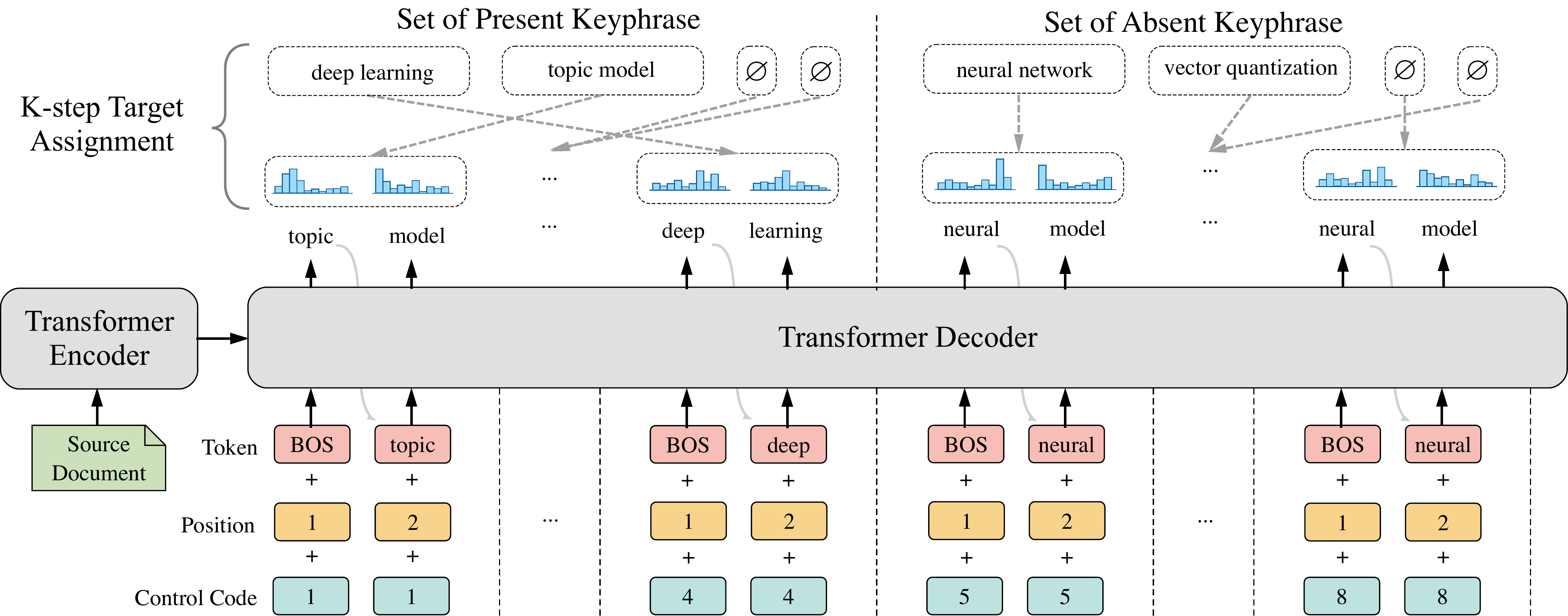}
\caption{Architecture of \textsc{SetTrans} model with $N$ learned control codes as input conditions. A $K$-step target assignment mechanism is used during training, where we first predict $K$ words for each code, and then find an optimal allocation among the predictions and targets. In the figure, $N=8$ and $K=2$ are used.}
\label{fig:model}
\end{figure*}

In previous CopyRNN based works, each source text corresponds to a single target keyphrase. Thus, the model needs beam search during inference to overgenerate multiple keyphrases, which cannot determine the dynamic number of keyphrases and consider the inter-relation among keyphrases. To this end, \citet{yuan2018} proposed an \textsc{One2Seq} training paradigm where each source text corresponds to a sequence of concatenated keyphrases. 
Thus, the model can capture the contextual information between the keyphrases as well as determines the dynamic number of keyphrases for different source texts. 
The recent works \citep{chan2019,chen2020a,swaminathan2020preliminary} mostly follow the \textsc{One2Seq} training paradigm. \citet{chan2019} proposed an RL-based fine-tuning method using $F_1$ and Recall metrics as rewards. \citet{swaminathan2020preliminary} proposed an RL-based fine-tuning method using a discriminator to produce rewards. All the above models need to be trained or pre-trained under the \textsc{One2Seq} paradigm. As keyphrases must be ordered before concatenating and keyphrases are inherently an unordered set, the model can be trained with wrong signal. Our \textsc{One2Set} training paradigm aims to solve this problem.

\section{Methodology}

This paper proposes a new training paradigm \textsc{One2Set} for keyphrase generation. A set prediction model based on Transformer (\textsc{SetTrans}) is proposed to fit this paradigm, as shown in Figure \ref{fig:model}. Given a fixed set of learned control codes as input conditions, the model generates a keyphrase or a special $\varnothing$ token for each code in parallel. During training, a $K$-step target assignment mechanism is proposed to dynamically determine the target corresponding to each code. The main idea is that the model first freely predicts $K$ steps without any supervision to see what keyphrase each code can roughly generate, and then use bipartite matching to find the optimal allocation based on the model's conjecture and target. Given the correspondence of each code and target, a separate set loss is then used to correct the model's conjecture, where half of the codes are trained to predict the present keyphrase set and the others are trained to predict the absent keyphrase set.

\subsection{The \textsc{One2Set} Training Paradigm}
We first formally describe the keyphrase generation task as follows. Given a document $\mathbf{x}$, it's aimed to predict a set of keyphrases $\mathcal{Y}=\{\mathbf{y}^i\}_{i=1,\dots,|\mathcal{Y}|}$, where $|\mathcal{Y}|$ is the number of keyphrases. To solve the KG task, previous works typically adopted an \textsc{One2One} training paradigm \citep{meng2017} or \textsc{One2Seq} training paradigm \citep{yuan2018}. The difference between the two training paradigms is that the form of training samples is different. Specifically, in the \textsc{One2One} training paradigm, each original sample pair $(\mathbf{x}, \mathcal{Y})$ is divided into multiple pairs $\{(\mathbf{x}, \mathbf{y}^i)\}_{i=1,\dots,|\mathcal{Y}|}$ to perform training independently. In the \textsc{One2Seq} training paradigm, each original sample pair is processed as $(\mathbf{x}, {f}(\mathcal{Y}))$, 
where ${f(\mathcal{Y})}$ is a sequence of keyphrases after the reordering and concatenating operation. 

To solve the wrong bias problem caused by the \textsc{One2Seq} training paradigm, we propose the \textsc{One2Set} training paradigm, where each original sample pair is kept still as $(\mathbf{x}, \mathcal{Y})$. Hence, the sample used in training is consistent with the original sample, which avoids the intractable problem introduced by the additional processing (i.e., dividing or concatenating).

\subsection{The \textsc{SetTrans} Model}
We adopt the Transformer \citep{vaswani2017attention} as the backbone encoder-decoder framework. 
However, the vanilla Transformer can only generate a sequence but not a set.
To predict a set of keyphrases, we propose \textsc{SetTrans} model that utilizes a set of learned control codes as additional decoder inputs. 
By performing generation conditioned on each control code, we can generate a set of keyphrases in parallel. 
To decide suitable numbers of keyphrases for different given documents, we fix the total length of the control codes to a sufficient number $N$, and introduce a special $\varnothing$ token that represents the meaning of ``no corresponding keyphrase''. Hence, we can determine the appropriate number of keyphrases for an input document after removing all the $\varnothing$ tokens from the $N$ predictions. 

Formally, the decoder input at time step $t$ for control code $n$ is defined as follows:
\begin{equation}
\mathbf{d}^n_{t} = \mathbf{e}^w_{y_{t-1}^n} + \mathbf{e}^p_t + \mathbf{c}^n,
\end{equation}
where $\mathbf{e}^w_{y_{t-1}^n}$ is the embedding of word $y_{t-1}^n$, $\mathbf{e}^p_t$ is the $t$-th sinusoid positional embedding as in \citep{vaswani2017attention} and $\mathbf{c}^n$ is the $n$-th learned control code embedding. The decoder outputs the predictive distribution $\mathbf{p}^n_{t}$, which is used to get the next word $y_{t}^n$ . 
As some keyphrases contain words that do not exist in the predefined vocabulary but appear in the input document, we also employ a copy mechanism \citep{see-etal-2017-get}, which is generally adopted for many previous KG works \citep{meng2017,chan2019,chen2020a,yuan2018}. 

\subsection{Training}
The main difficulty of training under the \textsc{One2Set} paradigm is that the correspondence between each prediction and ground-truth keyphrase is unknown, so that the cross-entropy loss cannot be directly used. Hence, we introduce a $K$-step target assignment mechanism to assign the ground-truth keyphrase for each prediction, and a separate set loss to train the model in an end-to-end way.

\subsubsection{$K$-step Target Assignment}
We first generate $K$ words for each control code and collect the corresponding predictive probability distributions of each step. Formally, we denote $\mathbf{P}=\{\mathbf{P}^n\}_{n=1,\dots,N}$, where $\mathbf{P}^n=\{\mathbf{p}^n_{t}\}_{t=1,\dots,K}$ 
and $\mathbf{p}^n_{t}$ is the predictive distribution at time step $t$ for control code $n$.

Then, we find a bipartite matching between the ground-truth keyphrases and predictions. Assuming the predefined number of control codes $N$ is larger than the number of ground-truth keyphrases, we consider the ground-truth keyphrases also as a set of size $N$ padded with $\varnothing$. Note that the bipartite matching enforces permutation-invariance, and guarantees that each target element has a unique match. Thus, it reduces the duplication ratio of predictions. Specifically, as shown in Figure \ref{fig:model}, both the fifth and eighth control code predict the same keyphrase ``neural model'', but one of them is assigned with $\varnothing$. The eighth code can perceive that this keyphrase has been generated by another code. Hence, the control codes can learn their mutual dependency during training and not generate duplicated keyphrases. 

Formally, to find a bipartite matching between sets of ground-truth keyphrases and predictions, we search for a permutation $\hat{\pi}$ with the lowest cost:
\begin{equation}
\hat{\pi}=\underset{\pi \in \Pi({N})}{\arg \min } \sum_{n=1}^{N} \mathcal{C}_{\operatorname{match}}\left(\mathbf{y}^{n}, \mathbf{P}^{\pi(n)}\right),
\end{equation}
where $\Pi(N)$ is the space of all $N$-length permutations, 
 $\mathcal{C}_{\operatorname{match}}\left(\mathbf{y}^{n}, \mathbf{P}^{\pi(n)}\right)$ is a pair-wise matching cost between the ground truth $\mathbf{y}^{n}$ and distributions of a prediction sequence with index $\pi(n)$. This optimal assignment is computed efficiently with the Hungarian algorithm \citep{kuhn1955hungarian}. The matching cost takes into account the class predictions, which can be defined as follows:
\begin{equation}
\begin{aligned}
\mathcal{C}_{\operatorname{match}}\left(\mathbf{y}^{n}, \mathbf{P}^{\pi(n)}\right) &= -\sum_{t=1}^{s} \mathbbm{1}_{\left\{y^n_{t} \neq  \varnothing\right\}} \mathbf{p}^{\pi(n)}_t\left(y^n_{t}\right),
\end{aligned}
\end{equation}
where $s=\min(|\mathbf{y}^{n}|, K)$ is the minimum shared length between the target and predicted sequence, $\mathbf{p}^{\pi(n)}_t\left(y^n_{t}\right)$ denotes the probability of word $y^n_{t}$ in $\mathbf{p}^{\pi(n)}_t$, and we ignore the score from matching predictions with $\varnothing$, which ensures that valid targets (i.e., non-$\varnothing$ targets) can be allocated to predictions with as higher predictive probability as possible.

\subsubsection{Separate Set Loss}
\label{sec:loss}

Given the correspondence between each code and target, we can train the model to predict a single target set, which is defined as follows:
\begin{equation}
\mathcal{L}(\theta)=-\sum_{n=1}^{N} \sum_{t=1}^{|\mathbf{y}^n|} \log \mathbf{\overline{p}}^{\hat{\pi}(n)}_t\left(y^n_{t}\right),
\label{eq:loss}
\end{equation}
where $\mathbf{\overline{p}}^{\hat{\pi}(n)}_t$ is the predictive probability distribution using teacher forcing.
However, predicting present and absent keyphrases requires the model to have different capabilities, we propose a separate set loss to flexibly take this bias into account in a unified model. Specifically, we first separate the control codes into two fixed sets with equal size of $N/2$, which is denoted as $\mathcal{C}_1$ and $\mathcal{C}_2$, and the target keyphrase set $\mathcal{Y}$ into present target keyphrase set $\mathcal{Y}^{pre}$ and absent target keyphrase set $\mathcal{Y}^{abs}$. Finally, the bipartite matching is performed on the two sets separately, namely, we find a permutation $\hat{\pi}^{pre}$ using $\mathcal{Y}^{pre}$ and predictions from $\mathcal{C}_1$, and $\hat{\pi}^{abs}$ using $\mathcal{Y}^{abs}$ and predictions from $\mathcal{C}_2$. Thus, we can modify the final loss in Equal \ref{eq:loss} as follows:
\begin{equation}
\begin{aligned}
\mathcal{L}(\theta)&=-(\sum_{n=1}^{N/2} \sum_{t=1}^{|\mathbf{y}^n|} \log \mathbf{\overline{p}}^{\hat{\pi}^{pre}(n)}_t\left(y^n_{t}\right)\\
&+\sum_{n=N/2+1}^{N} \sum_{t=1}^{|\mathbf{y}^n|} \log \mathbf{\overline{p}}^{\hat{\pi}^{abs}(n)}_t\left(y^n_{t}\right)).
\end{aligned}
\end{equation}
In practice, we down-weight the log-probability term when $y^n_{t} = \varnothing$ by scale factors $\lambda_{pre}$ and $\lambda_{abs}$ for present keyphrase set and absent keyphrase set to account for the class imbalance.

\section{Experimental Setup} 
\subsection{Datasets}
We conduct our experiments on five scientific article datasets, including \textbf{Inspec} \citep{hulth2003improved}, \textbf{NUS} \citep{nguyen2007keyphrase}, \textbf{Krapivin} \citep{krapivin2009large}, \textbf{SemEval} \citep{kim2010semeval} and \textbf{KP20k} \citep{meng2017}. 
Each sample from these datasets consists of a title, an abstract, and some keyphrases. Following previous works \citep{meng2017,chen2019a,chen2019,yuan2018}, we concatenate the title and abstract as a source document. 
We use the largest dataset (i.e., {KP20k}) to train all the models. After preprocessing (i.e., lowercasing, replacing all the digits with the symbol $\langle\mathit{digit}\rangle$ and removing the duplicated data), the final {KP20k} dataset contains 509,818 samples for training, 20,000 for validation, and 20,000 for testing. The dataset statistics are shown in Table
\ref{tab:datasets}.

\begin{table}[t]
\scalebox{0.73}{
\begin{tabular}{lcccc}
\toprule
 \textbf{Dataset} & \textbf{\#Samples} & \textbf{Avg. \#KP} & \textbf{Avg. $|$KP$|$} & \textbf{\% of Abs.KP} \\
 \hline
{Inspec} & 500 & 9.79 & 2.48 & 26.42 \\
{NUS} & 211 & 10.81 & 2.22 & 45.36 \\
{Krapivin} & 400 & 5.83 & 2.21 & 44.33 \\
{SemEval} & 100 & 14.43 & 2.38 & 55.61 \\
{KP20k} & 20,000 & 5.26 & 2.04 & 37.23 \\
\bottomrule
\end{tabular}}
\caption{Statistics of the testing set on five datasets. {\#KP}: number of keyphrases. {$|$KP$|$}: length of keyphrase. {Abs.KP}: absent keyphrases.}
\label{tab:datasets}
\end{table}

\subsection{Baselines}
We focus on the comparisons with the following state-of-the-art methods as our baselines:

\begin{compactitem}
\item \textbf{catSeq} \citep{yuan2018}. The RNN-based seq2seq model with copy mechanism trained under \textsc{One2Seq} paradigm. 

\item \textbf{catSeqTG} \citep{chen2019a}. An extension of catSeq with additional title encoding and cross-attention.

\item \textbf{catSeqTG-$2RF_1$} \citep{chan2019}. An extension of catSeqTG with RL-based fine-tuning using $F_1$ and Recall metrics as rewards.

\item \textbf{GAN$_{MR}$} \citep{swaminathan2020preliminary}. An extension of catSeq with RL-based fine-tuning using a discriminator to produce rewards.

\item \textbf{ExHiRD-h} \citep{chen2020a}. An extension of catSeq with a hierarchical decoding method and an exclusion mechanism to avoid generating duplicated keyphrases.
\end{compactitem}

In this paper, we propose two Transformer-based models that are denoted as follows:
\begin{compactitem}
\item \textbf{Transformer}. A Transformer-based model with copy mechanism trained under \textsc{One2Seq} paradigm.
\item \textbf{\textsc{SetTrans}}. An extension of Transformer with additional control codes trained under \textsc{One2Set} paradigm.
\end{compactitem}

\begin{table*}[htbp]
\centering
\scalebox{0.7}{
\begin{tabular}{l|ll|ll|ll|ll|ll}
\toprule
\multicolumn{1}{l|}{\multirow{2}{*}{\textbf{Model}}} & \multicolumn{2}{c|}{\textbf{Inspec}} & \multicolumn{2}{c|}{\textbf{NUS}} & \multicolumn{2}{c|}{\textbf{Krapivin}} & \multicolumn{2}{c|}{\textbf{SemEval}} & \multicolumn{2}{c}{\textbf{KP20k}} \\
 & {$F_1@5$} & {$F_1@M$} & {$F_1@5$} & {$F_1@M$} & {$F_1@5$} & {$F_1@M$} & {$F_1@5$} & {$F_1@M$} & {$F_1@5$} & {$F_1@M$} \\ 
\hline
catSeq \citep{yuan2018} & 0.225 & 0.262 & 0.323 & 0.397 & 0.269 & 0.354 & 0.242 & 0.283 & 0.291 & 0.367 \\
catSeqTG \citep{chen2019a} & 0.229 & 0.270 & 0.325 & 0.393 & 0.282 & 0.366 & 0.246 & 0.290 & 0.292 & 0.366 \\
catSeqTG-$2RF_1$ \citep{chan2019} & 0.253 & 0.301 & 0.375 & 0.433 & 0.300 & \textbf{0.369} & 0.287 & 0.329 & 0.321 & 0.386 \\
GAN$_{MR}$ \citep{swaminathan2020preliminary} & 0.258 & 0.299 & 0.348 & 0.417 & 0.288 & \textbf{0.369} & - & - & 0.303 & 0.378 \\
ExHiRD-h \citep{chen2020a} & 0.253 & 0.291 & - & - & 0.286 & 0.347 & 0.284 & 0.335 & 0.311 & 0.374 \\
\hline
Transformer (\textsc{One2Seq})& 0.281$_5$ & \textbf{0.325$_6$} & 0.370$_7$ & 0.419$_{10}$ & 0.315$_8$ & {0.365$_{5}$} & 0.287$_{14}$ & 0.325$_{15}$ & 0.332$_1$ & 0.377$_1$ \\
\textsc{SetTrans} (\textsc{One2Set}) & \textbf{0.285$_3$} & {0.324$_3$} & \textbf{0.406$_{12}$} & \textbf{0.450$_7$} & \textbf{0.326$_{12}$} & {0.364$_{12}$} & \textbf{0.331$_{20}$} & \textbf{0.357$_{13}$} & \textbf{0.358$_5$} & \textbf{0.392$_4$} \\
\bottomrule
\end{tabular}}
\caption{Present keyphrases prediction results of all models. The best results are bold. The subscript represents the corresponding standard deviation (e.g., 0.392$_4$ indicates 0.392$\pm$0.004).}
\label{tab:present}
\end{table*}

\begin{table*}[htbp]
\centering
\scalebox{0.7}{
\begin{tabular}{l|ll|ll|ll|ll|ll}
\toprule
\multicolumn{1}{l|}{\multirow{2}{*}{\textbf{Model}}} & \multicolumn{2}{c|}{\textbf{Inspec}} & \multicolumn{2}{c|}{\textbf{NUS}} & \multicolumn{2}{c|}{\textbf{Krapivin}} & \multicolumn{2}{c|}{\textbf{SemEval}} & \multicolumn{2}{c}{\textbf{KP20k}} \\
 & {$F_1@5$} & {$F_1@M$} & {$F_1@5$} & {$F_1@M$} & {$F_1@5$} & {$F_1@M$} & {$F_1@5$} & {$F_1@M$} & {$F_1@5$} & {$F_1@M$} \\ 
\hline
catSeq \citep{yuan2018} & 0.004 & 0.008 & 0.016 & 0.028 & 0.018 & 0.036 & 0.016 & 0.028 & 0.015 & 0.032 \\
catSeqTG \citep{chen2019a} & 0.005 & 0.011 & 0.011 & 0.018 & 0.018 & 0.034 & 0.011 & 0.018 & 0.015 & 0.032 \\
catSeqTG-$2RF_1$ \citep{chan2019} & 0.012 & 0.021 & 0.019 & 0.031 & 0.030 & 0.053 & 0.021 & {0.030} & 0.027 & 0.050 \\
GAN$_{MR}$ \citep{swaminathan2020preliminary} & 0.013 & 0.019 & 0.026 & 0.038 & 0.042 & 0.057 & - & - & 0.032 & 0.045 \\
ExHiRD-h \citep{chen2020a} & 0.011 & 0.022 & - & - & 0.022 & 0.043 & 0.017 & 0.025 & 0.016 & 0.032 \\
\hline
Transformer (\textsc{One2Seq})& 0.010$_{2}$ & 0.019$_{4}$ & 0.028$_{2}$ & 0.048$_{2}$ & 0.032$_{1}$ & 0.060$_{4}$ & 0.020$_{5}$ & 0.023$_{3}$ & 0.023$_{1}$ & 0.046$_{1}$ \\
\textsc{SetTrans} (\textsc{One2Set})& \textbf{0.021$_{1}$} & \textbf{0.034$_{3}$} & \textbf{0.042$_{2}$} & \textbf{0.060$_{4}$} & \textbf{0.047$_{7}$} & \textbf{0.073$_{11}$} & \textbf{0.026$_{3}$} & \textbf{0.034$_{5}$} & \textbf{0.036$_{2}$} & \textbf{0.058$_{3}$} \\
\bottomrule
\end{tabular}}
\caption{Absent keyphrases prediction results of all models. 
The best results are bold. The subscript represents the corresponding standard deviation (e.g., 0.058$_3$ indicates 0.058$\pm$0.003).
}
\label{tab:absent}
\end{table*}

\subsection{Implementation Details}
Following previous works \citep{chan2019,chen2020a,yuan2018}, when training under the \textsc{One2Seq} paradigm, the target keyphrase sequence is the concatenation of present and absent keyphrases, with the present keyphrases are sorted according to the orders of their first occurrences in the document and the absent keyphrase kept in their original order.
We use a Transformer structure similar to \citet{vaswani2017attention}, with six layers and eight self-attention heads, 2048 dimensions for hidden states. In the training stage, we choose the top 50,002 frequent words to form the predefined vocabulary and set the embedding dimension to 512. We use the Adam optimization algorithm \citep{kingma2014adam} with a learning rate of 0.0001, and a batch size of 12.
During testing, we use greedy search as the decoding algorithm. We set the number of control codes to 20 as we find it covers 99.5\% of the samples in the validation set. We use a number of two for target assignment steps $K$ based on the average keyphrase length on the validation set, a factor of 0.2 and 0.1 for $\lambda_{pre}$ and $\lambda_{abs}$ respectively based on the validation set.
We conduct the experiments on a GeForce RTX 2080Ti GPU, repeat three times using different random seeds, and report the averaged results.

\subsection{Evaluation Metrics}
We follow previous works \citep{chan2019,chen2020a} and use macro-averaged $F_1@5$ and $F_1@M$ for both present and absent keyphrase predictions. $F_1@M$ compares all the keyphrases predicted by the model with the ground-truth keyphrases, which means it considers the number of predictions. 
For $F_1@5$, when the prediction number is less than five, we randomly append incorrect keyphrases until it obtains five predictions. If we do not adopt such an appending operation, $F_1@5$ will become the same with $F_1@M$ when the prediction number is less than five as shown in \citet{chan2019}. 
We apply the Porter Stemmer before determining whether two keyphrases are identical and remove all the duplicated keyphrases after stemming. 

\section{Results and Analysis}
\subsection{Present and Absent Keyphrase Predictions}

Table \ref{tab:present} and Table \ref{tab:absent} show the performance evaluations of the present and absent keyphrase, respectively. 
We observe that the proposed \textsc{SetTrans} model consistently outperforms almost all the previous state-of-the-art models on both $F_1@5$ and $F_1@M$ metrics by a large margin, which demonstrates the effectiveness of our methods. As noted by previous works \citep{chan2019,yuan2018} that predicting absent keyphrases for a document is an extremely challenging task, thus the performance is much lower than that of present keyphrase prediction. 
Regarding the comparison of our Transformer model trained under \textsc{One2Seq} paradigm and \textsc{SetTrans} model trained under \textsc{One2Set} paradigm, we find \textsc{SetTrans} model consistently improves both keyphrase extractive and generative ability by a large margin on almost all the datasets, and maintains the performance of present keyphrase prediction on the Inspec and Krapivin datasets, which demonstrates the advantages of \textsc{One2Set} training paradigm.

\subsection{Diversity of Predicted Keyphrases}
To investigate the model's ability to generate diverse keyphrases, we measure the average numbers of unique present and absent keyphrases, and the average duplication ratio of all the predicted keyphrases. The results are reported in Table \ref{tab:dup}. Based on the results, we observe that our \textsc{SetTrans} model generates more unique keyphrases than other baselines by a large margin, as well as achieves a significantly lower duplication ratio. Note that ExHiRD-h specifically designed a deduplication mechanism to remove duplication in the inference stage. In contrast, our model achieves a lower duplication ratio without any deduplication mechanism, which proves its effectiveness. However, we also observe that our model tends to overgenerate more present keyphrases than the ground-truth on the {Krapivin} and {KP20k} datasets. We analyze that different datasets have different preferences for the number of keyphrases, which we leave as our future work.

\begin{table}[t]
\centering
\scalebox{0.58}{
\begin{tabular}{l|ccc|ccc|ccc}
\toprule
\multicolumn{1}{l|}{\multirow{2}{*}{\textbf{Model}}} & \multicolumn{3}{c|}{\textbf{Krapivin}} & \multicolumn{3}{c|}{\textbf{SemEval}} & \multicolumn{3}{c}{\textbf{KP20k}} \\
\multicolumn{1}{c|}{} & \#PK & \#AK & Dup & \#PK & \#AK & Dup & {\#PK} & \#AK & Dup \\
\hline
Oracle & 3.24 & 2.59 & - & 6.12 & 8.31 & - & 3.31 & 1.95 & - \\
\hline
catSeq & 3.50 & 0.67 & 0.46 & 3.48 & 0.77 & 0.53 & 3.71 & 0.55 & 0.39 \\
catSeqTG & {3.82} & {0.83} & 0.41 & 3.82 & 1.09 & {0.63} & 3.77 & 0.67 & 0.36 \\
catSeqTG-$2RF_1$ & \textbf{3.28} & 1.56 & 0.29 & 3.57 & 1.50 & {0.25} & \textbf{3.55} & 1.44 & 0.28 \\
ExHiRD-h & {4.41} & {1.02} & 0.14 & 3.65 & 0.99 & 0.09 & 3.97 & 0.81 & 0.11 \\
\hline
Transformer & {4.44} & {1.39} & 0.29 & 4.30 & 1.52 & 0.27 & 4.64 & 1.16 & 0.26 \\
\textsc{SetTrans} & {4.83} & \textbf{2.20} & \textbf{0.08} & \textbf{4.62} & \textbf{2.18} & \textbf{0.08} & {5.10} & \textbf{2.01} & \textbf{0.08} \\
\bottomrule
\end{tabular}}
\caption{Number and duplication ratio of predicted keyphrases on three datasets. ``\#PK'' and ``\#AK'' are the average number of unique present and absent keyphrases respectively. ``Dup'' refers to the average duplication ratio of predicted keyphrases. ``Oracle'' refers to the gold average keyphrase number. }
\label{tab:dup}
\end{table}

\subsection{Ablation Study}
\label{sec:ablation}

To understand the effects of each component of the \textsc{SetTrans} model, we conduct an ablation study on it and report the results on the {KP20k} dataset in Table \ref{tab:ablation}. 

\paragraph{Effects of Model Architecture}
To verify the effectiveness of the model architecture of \textsc{SetTrans}, we remove the control codes and find the model is completely broken. The duplication ratio increases to 0.95, which means all the 20 control codes predict the same keyphrase. This occurs because when the control codes are removed, all the predictions depend on the same condition (i.e., the source document) without any distinction. This demonstrates that the control codes play an extremely important role in the \textsc{SetTrans} model.

\paragraph{Effects of Target Assignment}
The major difficulty for successfully training under \textsc{One2Set} paradigm is the target assignment between predictions and targets. An attempt is first made to remove the $K$-step target assignment mechanism, which means that we employ a fixed sequential matching strategy as in the \textsc{One2Seq} paradigm. From the results, we observe that both the present and absent keyphrase performances degrade, the number of predicted keyphrases also drops dramatically, and the duplication ratio increased greatly by 18\%. We analyze the reasons as follows: (1) {The dynamic characteristics of the $K$-step target assignment remove unnecessary position constraint during training, which encourages the model to generate more keyphrases.} Specifically, the model can generate a keyphrase in any location rather than only in the given position. Thus, the model does not need to consider the position constraint during the generation and encourages all the control codes to predict keyphrases rather than only the first few codes, which will be verified in Section \ref{sec:code}.
(2) {The bipartite characteristics of the $K$-step target assignment forces the model to predict unique keyphrases, which reduces the duplication ratio of predictions.} When predictions from two codes are similar, only one code may be assigned a target keyphrase, and the other is assigned a $\varnothing$ token. Thus, the model can be very careful about each prediction to prevent duplication. 
We further experiment that replacing the $K$-step target assignment with a random assignment, and we find that the results are similar to those when removing the control codes. This is because the random assignment misleads the learning of the control codes and causes them to become invalid.

\begin{table}[t]
\centering
\scalebox{0.6}{
\begin{tabular}{l|ccc|ccc|c}
\toprule
\multirow{2}{*}{\textbf{Model}} & \multicolumn{3}{c|}{\textbf{Present}} & \multicolumn{3}{c|}{\textbf{Absent}} & \multirow{2}{*}{\textbf{Dup}} \\
 & {$F_1@5$} & {$F_1@M$} & \#PK & {$F_1@5$} & {$F_1@M$} & \#AK &  \\
\hline
Oracle & - & - & 3.31 & - & - & 1.95 & - \\
\hline
\textsc{SetTrans} & \textbf{0.358} & \textbf{0.392} & 5.10 & \textbf{0.036} & \textbf{0.058} & \textbf{2.01} & 0.08 \\
\hline
\textit{Model Architecture} &  &  &  &  &  &  &  \\
- control codes & {0.001} & {0.002} & 0.01 & 0.000 & 0.000 & 0.00 & 0.95 \\
\hline
\textit{{Target Assignment}} &  &  &  &  &  &  &  \\
- $K$-step assign & 0.265 & 0.381 & \textbf{2.64} & 0.020 & 0.045 & 0.81 & 0.26 \\
\, + random assign & 0.005 & 0.010 & 1.05 & 0.001 & 0.002 & 0.04 & 0.95 \\
\hline
\textit{{Set Loss}} & \textbf{} & \textbf{} &  & \textbf{} & \textbf{} & \textbf{} & \textbf{} \\
- teacher forcing & 0.001 & 0.002 & 0.01 & 0.000 & 0.000 & 0.00 & 0.89 \\
- separate set loss & 0.355 & 0.383 & 5.31 & 0.016 & 0.031 & 0.55 & \textbf{0.05} \\
\bottomrule
\end{tabular}}
\caption{Ablation study of \textsc{SetTrans} on {KP20k} dataset. ``- teacher forcing'' refers to directly calculating the loss after target assignment in a student forcing schema. ``- separate set loss'' refers to using a single set loss. 
}
\label{tab:ablation}
\end{table}

\paragraph{Effects of Set Loss}
As discussed in Section \ref{sec:loss}, teacher forcing and a separate set loss are used to train the model after assigning a target for each prediction. We investigate their effects in detail. The results show the following. (1) {Teaching forcing can alleviate the cold start problem.} After removing teaching forcing, the model faces a cold start problem, in other words, the lack of supervision information leads to a poor prediction, and the target assignment is therefore not ideal, which causes the model to fail at the early stage of training. (2) {A separate set loss helps in both present and absent keyphrase predictions but also increases the duplication ratio slightly compared with a single set loss.} As producing correct present keyphrases is an easier task, the model tends to generate present keyphrases only when using a single set loss. Our separate set loss can infuse different inductive biases into the two sets of control codes, which makes them more focused on generating one type of keyphrase (i.e., the present one or absent one). Thus, it increases the accuracy of the predictions and encourages more absent keyphrase predictions. However, because bipartite matching is performed separately, the constraint of unique prediction does not exist between the two sets, which leads to a slight increase in the duplication ratio.

\subsection{Performance over Scale Factors}
\label{sec:scale}
\begin{figure}[t]
\centering
\includegraphics[width=3in]{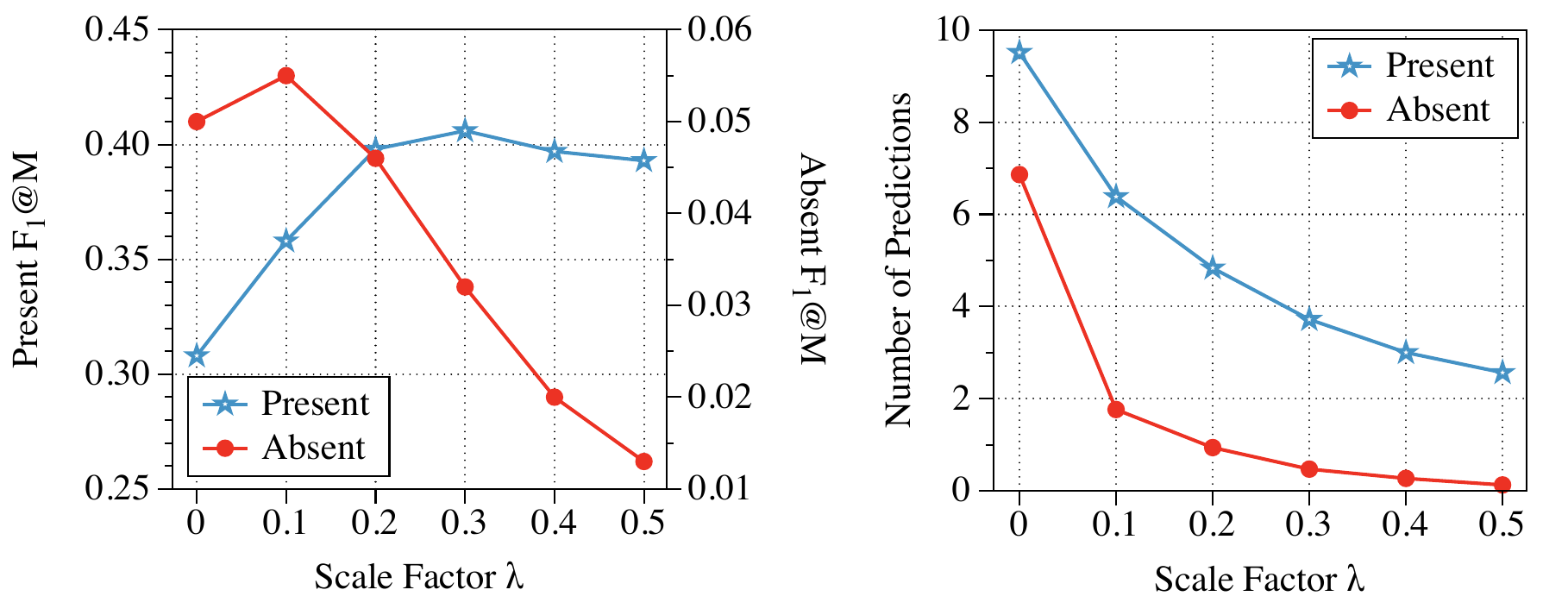}
\caption{Performance and number of predictions for present and absent keyphrase under different loss scale factors $\lambda$ for $\varnothing$ token on {KP20k} dataset. We set both $\lambda_{pre}$ and $\lambda_{abs}$ to $\lambda$ to simplify the comparison.}
\label{fig:combine}
\end{figure}

In this section, we conduct experiments on {KP20k} dataset to evaluate performance under different loss scale factors $\lambda$ for $\varnothing$ token. The results are shown in Figure \ref{fig:combine}. 

The left part of the figure shows that when $\lambda=0.2$, the performances on both present and absent keyphrases are consistently better than the results when $\lambda=0.1$. However, a scale factor larger than 0.1 improves the present keyphrase performance, but also harms the absent keyphrase performance. As we can see from the right part of the figure, the number of predictions decreases consistently for both the present and absent keyphrases when the scale factor becomes larger. This is because a larger scale factor causes the model to predict more $\varnothing$ tokens to reduce the loss penalty during training. Moreover, we also find that the precision metric $P@M$ will increases when the number of predictions decreases. While the effect of the decrease in the recall metric $R@M$ is even greater when the number is too small, which leads to a degradation in the overall metric $F_1@M$.

\subsection{Efficiency over Assignment Steps}
In this section, we study the influence of target assignment steps $K$ on the prediction performance and efficiency compared with Transformer. 

\begin{figure}[t]
\centering
\includegraphics[width=3in]{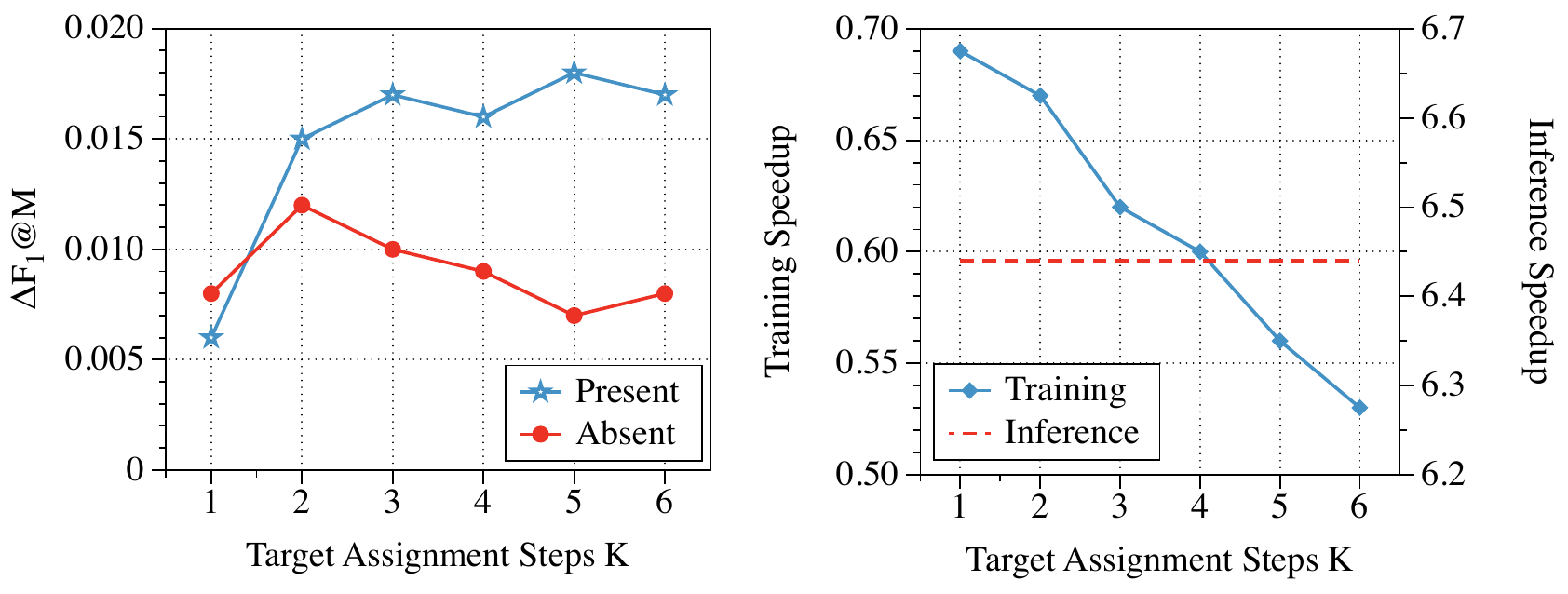}
\caption{Performance and training/inference speedup compared with Transformer over different target assignment steps $K$ on KP20k dataset. }
\label{fig:combine_k}
\end{figure}

As shown in the left part of Figure \ref{fig:combine_k}, we note that when $K$ is equal to 1, the improvement of \textsc{SetTrans} over Transformer is relatively lower than when it is equal to 2 (i.e., the average length of keyphrase). This is mainly because some keyphrases that have the same first word cannot be distinguished during training, which could interfere with the learning of control codes. The right part of Figure \ref{fig:combine_k} shows the training and inference speedup with various $K$ compared with the Transformer.
We note \textsc{SetTrans} could be slower than Transformer at the training stage, and a smaller $K$ could alleviate this problem. For performance and efficiency considerations, we consider 2 to be an appropriate value for steps $K$. Moreover, as $K$ is only used in the training stage, \textsc{SetTrans} is 6.44 times invariably faster than Transformer on the inference stage. This is because that with different control codes as input condition, all the keyphrases can be generated in parallel on the GPU. Hence, in addition to better performance than Transformer, \textsc{SetTrans} also has great advantages in the inference efficiency.

% \begin{table}[t]
% \centering
% \scalebox{0.8}{
% \begin{tabular}{l|cc}
% \toprule
% \textbf{} & \textbf{Training} & \textbf{Inference} \\
% \hline
% Transformer & 1x & 1x \\
% \textsc{SetTrans} & 0.67x & 6.44x \\
% \bottomrule
% \end{tabular}}
% \caption{Comparison of training and inference speed for Transformer under \textsc{One2Seq} paradigm and \textsc{SetTrans} under \textsc{One2Set} paradigm.}
% \label{tab:speed}
% \end{table}

\begin{figure}[t]
\centering
\includegraphics[width=3in]{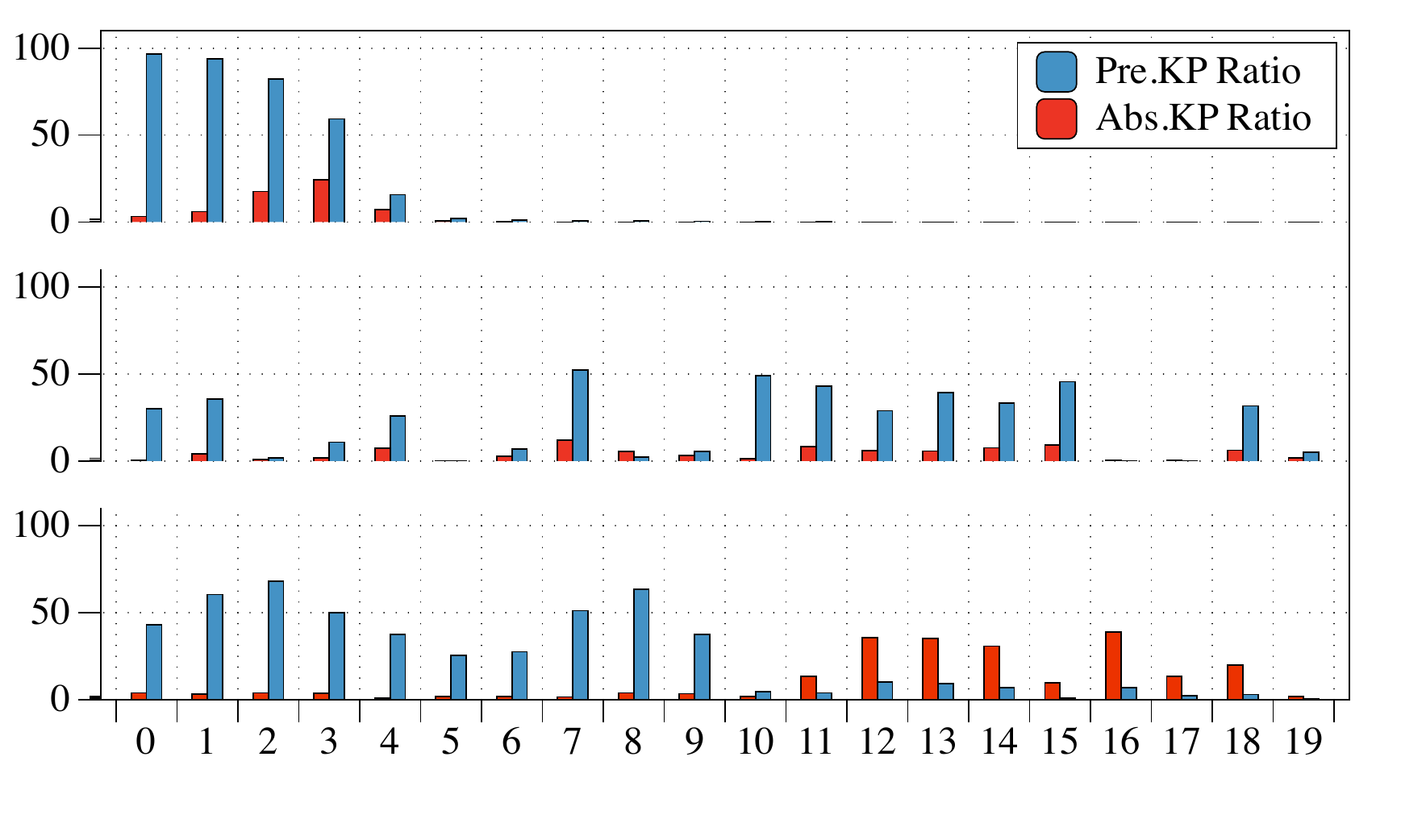}
\caption{Ratio of present and absent keyphrase predictions for all the control codes on {KP20k} dataset. The subgraphs from top to bottom are for the ``w/o $K$-step target assignment'', ``a single set loss'', and ``a separate set loss'' cases, respectively. The summation of the ratios of the present keyphrases, absent keyphrases and $\varnothing$ equals to 100\% for each code. }
\label{fig:codes}
\end{figure}

\subsection{Analysis of Learned Control Codes}
\label{sec:code}

Our analysis here is driven by two questions from Section \ref{sec:ablation}:

(1) Whether the $K$-step target assignment mechanism encourages all the control codes to predict keyphrases rather than only the first few codes? 

(2) Whether the separate set loss makes the control codes more focused on generating one type of keyphrase (i.e., present or absent) compared to the single set loss? 

To investigate these two questions, we measure the ratio of present and absent keyphrase predictions for all the control codes on the {KP20k} dataset, which is shown in Figure \ref{fig:codes}. As shown in the top and middle subfigures, we observe that without the target assignment mechanism, many control codes are invalid (i.e., only predicting $\varnothing$), and only the first small part performs valid predictions. Moreover, when there are already very few valid predictions, the model still has a duplication ratio of up to 26\%, as shown in Table \ref{tab:ablation}, resulting in an even smaller number of final predictions. After the introduction of the target assignment mechanism, most of the codes can generate valid keyphrases, which increases the number of predictions. 

However, as shown in the middle subfigure, most of the control code tends to generate more present keyphrases than absent keyphrases when using a single set loss. When using a separate set loss in the bottom subfigure, the two parts are more inclined to predict only present and absent keyphrases respectively, which also increases the number of absent keyphrase predictions.

\section{Conclusions}
In this paper, we propose a new training paradigm \textsc{One2Set} without predefining an order to concatenate the keyphrases, and a novel model \textsc{SetTrans} that predicts a set of keyphrases in parallel. To successfully train under \textsc{One2Set} paradigm, we propose a $K$-step target assignment mechanism and a separate set loss, which greatly increases the number and diversity of the generated keyphrases. Experiments show that our method gains significantly huge performance improvements against existing state-of-the-art models. We also show that \textsc{SetTrans} has great advantages in the inference efficiency compared with the Transformer under \textsc{One2Seq} paradigm.

\section*{Acknowledgments}
The authors wish to thank the anonymous reviewers for their helpful comments. This work was partially funded by China National Key R\&D Program (No. 2017YFB1002104), National Natural Science Foundation of China (No. 62076069, 61976056), Shanghai Municipal Science and Technology Major Project (No.2021SHZDZX0103).

\bibliographystyle{acl_natbib}
\bibliography{acl2021}

\end{document}